\documentclass[sigconf,nonacm]{acmart}
\AtBeginDocument{%
  }

\setcopyright{acmlicensed}
\copyrightyear{2018}
\acmYear{2018}
\acmDOI{XXXXXXX.XXXXXXX}
\acmConference[Conference acronym 'XX]{Make sure to enter the correct
  conference title from your rights confirmation email}{June 03--05,
  2018}{Woodstock, NY}
\acmISBN{978-1-4503-XXXX-X/2018/06}

\usepackage{multirow}
\usepackage{enumitem}
\usepackage{amsthm}
\usepackage{fancyvrb}
\usepackage{booktabs}
\usepackage{hyperref}
\usepackage{fontawesome5}
\usepackage{academicons}
\usepackage{fvextra}
\usepackage[most]{tcolorbox}
\usepackage[table]{xcolor}
\definecolor{rowgray}{gray}{0.95}

\theoremstyle{definition} 
\newtheorem{defn}{Definition}[section] 

\begin{document}


\title{Sparking Scientific Creativity via\protect\\LLM-Driven Interdisciplinary Inspiration}

\author{Priyanka Kargupta, Shuhaib Mehri, Dilek Hakkani-Tur, Jiawei Han}

\affiliation{%
  \department{Siebel School of Computing and Data Science}
  \institution{University of Illinois at Urbana-Champaign}
  \city{Urbana, IL}
  \country{United States}
}

\email{pk36, mehri2, dilek, hanj@illinois.edu}

\renewcommand{\shortauthors}{Kargupta et al.}

\begin{abstract}
Despite interdisciplinary research leading to larger and longer-term impact, most work remains confined to single-domain academic silos. Recent AI-based approaches to scientific discovery show promise for interdisciplinary research, but many prioritize rapidly designing experiments and solutions, bypassing the exploratory, collaborative reasoning processes that drive creative interdisciplinary breakthroughs. As a result, prior efforts largely prioritize \textit{automating} scientific discovery rather than augmenting the reasoning processes that underlie scientific disruption. We present \textbf{\textsc{Idea-Catalyst}}, a novel framework that systematically identifies interdisciplinary insights to support creative reasoning in both humans and large language models. Starting from an abstract research goal, \textbf{\textsc{Idea-Catalyst}} is designed to assist the brainstorming stage, explicitly avoiding premature anchoring on specific solutions. The framework embodies key metacognitive features of interdisciplinary reasoning: \textbf{\textit{(a)}} defining and assessing research goals, \textbf{\textit{(b)}} awareness of a domain’s opportunities and unresolved challenges, and \textbf{\textit{(c)}} strategic exploration of interdisciplinary ideas based on impact potential. Concretely, \textbf{\textsc{Idea-Catalyst}} decomposes an abstract goal (e.g., improving human-AI collaboration) into core target-domain research questions that guide the analysis of progress and open challenges within that domain. These challenges are reformulated as domain-agnostic conceptual problems, enabling retrieval from external disciplines (e.g., Psychology, Sociology) that address analogous issues. By synthesizing and recontextualizing insights from these domains back into the target domain, \textbf{\textsc{Idea-Catalyst}} ranks source domains by their interdisciplinary potential. Empirically, this targeted integration improves average novelty by \textbf{21\%} and insightfulness by \textbf{16\%}, while remaining grounded in the original research problem. Overall, \textbf{\textsc{Idea-Catalyst}} provides a structured framework for boundary-spanning research ideation, with implications for both AI-assisted human creativity and automated scientific discovery.
\vspace{-1mm}
\begin{center}
\href{http://pkargupta.github.io/idea_catalyst.html}{\faGlobe\ \textbf{Project Page}} \quad
\href{https://github.com/pkargupta/idea_catalyst}{\faGithub\ \textbf{Repository}} \quad
\href{https://huggingface.co/datasets/pkargupta/idea_catalyst}{\faDatabase\ \textbf{Dataset}}
\end{center}
\end{abstract}

\ccsdesc[500]{Computing methodologies~Knowledge representation and reasoning}

\keywords{Scientific discovery, creative reasoning, human-AI collaboration}

\received{20 February 2007}
\received[revised]{12 March 2009}
\received[accepted]{5 June 2009}


\maketitle

\section{Introduction}

\par
\begin{figure}[h]
  \includegraphics[width=0.48\textwidth]{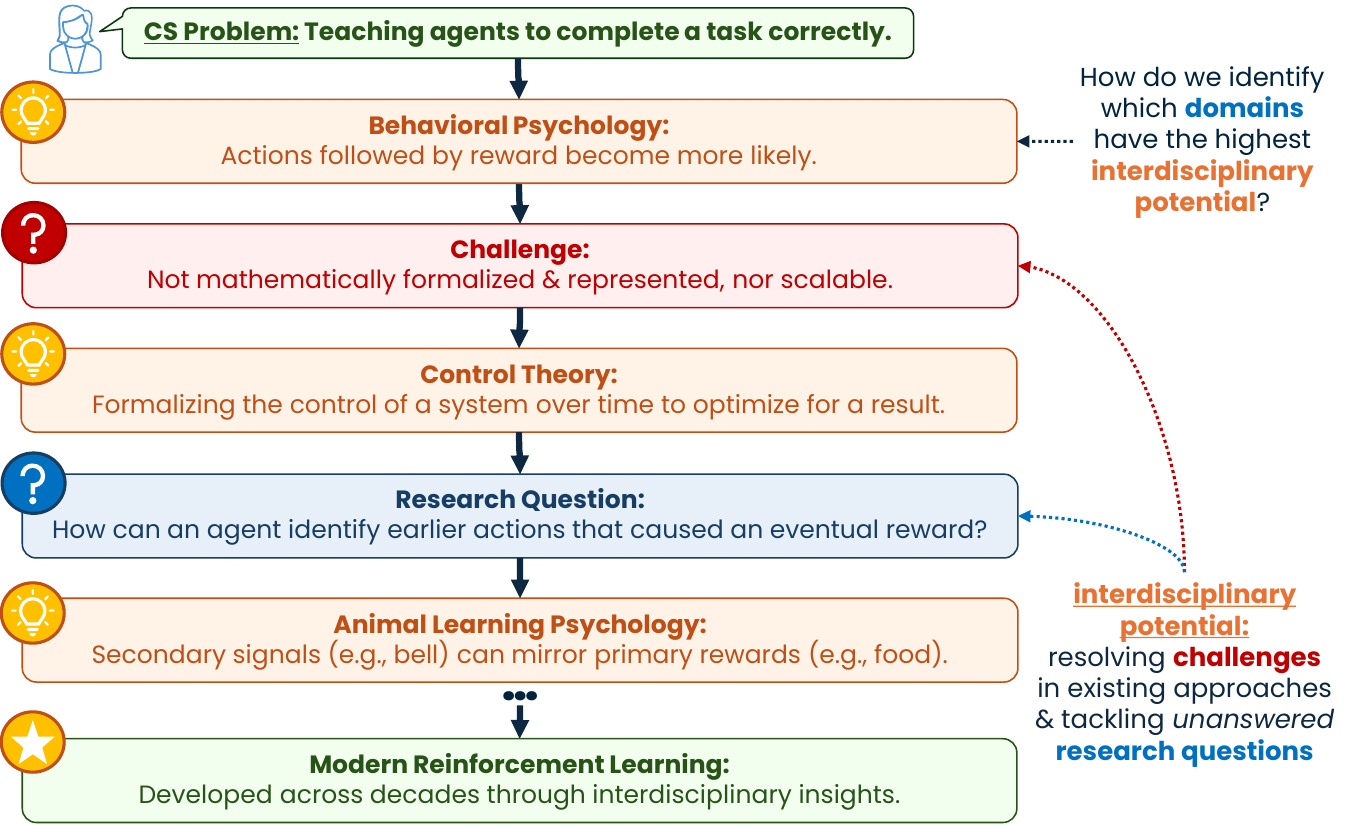}
  \caption[figure]{Interdisciplinary process of formalizing RL \cite{sutton1998reinforcement}.}
  \label{fig:teaser}
\end{figure}

 Scientific breakthroughs rarely arise from a single, isolated ``\textit{eureka}'' moment. Instead, research ideation unfolds as a complex, iterative formation process in which creative solutions emerge through the gradual synthesis of many fragmentary and partial ideas~\cite{sosa2019accretion,gonccalves2021life}. These early, tangible conceptual fragments, often originating from multiple domains~\cite{yanai2019night}, act as seeds for discussion, critique, and collaboration. They open up more meaningful research questions and address each other's conceptual gaps and challenges, gradually coalescing into more mature research directions ~\cite{sosa2019accretion,gonccalves2021life}. This pattern recurs throughout the history of scientific progress. Reinforcement learning, now a foundational paradigm in machine learning, did not originate within a single field but instead emerged from the convergence of behavioral psychology’s reward-driven learning principles, control theory’s mathematical formalizations, and animal learning psychology’s insights into secondary reinforcement signals~\cite{sutton1998reinforcement}, as illustrated in Figure~\ref{fig:teaser}. The field advanced through the accumulation, recombination, and refinement of ideas contributed by researchers operating under diverse conceptual lenses.

\par Empirical evidence shows that interdisciplinary synthesis yields substantially higher long-term impact, with each additional discipline increasing citation impact by approximately 20\% \cite{van2015interdisciplinary,okamura2019interdisciplinarity}. Yet despite this benefit, deeply integrative interdisciplinary research remains rare and fragile: only 5\% of cross-domain work involves high-involvement collaboration across non-neighboring fields \cite{porter2009science, raasch2013rise}. A central challenge, then, is how to \textit{foster interdisciplinary scientific creativity} and help researchers move beyond their academic silos. Recent work on AI-driven scientific discovery has explored the notion of ``AI co-scientists,'' in which large language models (LLMs) support (and in many cases automate) key stages of the research process, including ideation, experimentation, and critique \cite{gottweis2025towards, goel2025training, si2026towards, jansen2025codescientist}. Prior studies \cite{si2024can} comparing human- and LLM-generated ideas highlight a complementary but unresolved tension~\cite{si2024can}. Human-generated ideas are typically well grounded in existing literature, datasets, and practical constraints, but tend to remain focused on familiar problem formulations within a single domain. LLM-generated ideas, in contrast, exhibit a stronger tendency to draw inspiration from other disciplines, yet often do so in surface-level or stereotyped ways that undermine motivation, feasibility, and practical grounding \cite{si2024can, si2026towards, gupta2025all}.

\par Attempts to address these limitations by tightly coupling LLM ideation with automatic experiment execution \cite{si2026towards,jansen2025codescientist} introduce further trade-offs. While grounding ideas through execution improves feasibility, it also drives convergence toward incremental, single-domain refinements, eroding the exploratory strengths and cross-domain potential that initially distinguish LLM-based ideation \cite{si2026towards}. Similarly, other works primarily focus on generating execution plans from high-level research ideas, bypassing the exploratory and collaborative processes of ideation \cite{goel2025training, huang2025idea2plan}. This pattern reflects a broader challenge in research ideation: early-stage evaluation, especially when driven by empirical validation, can be counterproductive for creativity. As prior work notes, premature data and evaluation can ``cut the conversation'' and curtail exploration of the broader space of possibilities \cite{bose2013enhancing, catmull2023creativity}.

\par To avoid prematurely anchoring on end-to-end automated solutions and instead augment the initial creative ideation process itself, we propose \textbf{\textsc{Idea-Catalyst}}, a novel framework for the automatic generation of insightful, interdisciplinary idea fragments. Given a research problem in a target domain (e.g., ``\textit{teaching agents to complete a task}'' in Computer Science), our goal is to surface conceptual insights from external ``source'' domains (e.g., \textit{Behavioral Psychology}, \textit{Control Theory}, \textit{Animal Learning Psychology}) that can either help resolve persistent challenges in existing target-domain approaches or address research questions that remain unanswered within the target domain. Our retrieval-augmented, hierarchical ideation framework is guided by three core principles:
\vspace{-1mm}
\begin{enumerate}[leftmargin=5mm]
    \item \textit{\textbf{Analyzing the target domain to assess progress and reveal challenges.}} We first decompose the overarching research problem into a structured set of core research questions. By retrieving and analyzing target-domain literature conditioned on each question, the framework identifies what has already been addressed, where progress is uneven, and which challenges remain unresolved. Crucially, this analysis distinguishes between \textit{domain-specific challenges} (e.g., getting an algorithm to learn reliably from limited or noisy feedback) and deeper \textit{conceptual challenges} (e.g., understanding what an agent should aim for when its goals or feedback are unclear or change over time) that persist despite extensive prior work.
    \item \textit{\textbf{Exploring external source domains to uncover conceptually analogous solutions.}} For each unresolved conceptual challenge, the framework queries \& analyzes a diverse set of external domains to determine whether similar problems have been previously studied or solved under different assumptions, formalisms, or empirical settings. This step emphasizes cross-domain awareness, enabling the discovery of alternative perspectives, mechanisms, and abstractions that are absent from the target domain but potentially transferable.
    \item \textit{\textbf{Recontextualizing and strategically prioritizing interdisciplinary insights.}} Finally, extracted insights from relevant source domains are recontextualized into the language and constraints of the target domain, forming candidate idea fragments. These fragments are then ranked based on their potential to address high-impact challenges, balancing conceptual novelty with relevance to the original research goals. This strategic prioritization supports exploratory ideation while avoiding premature convergence or feasibility-driven pruning.
\end{enumerate}

\par Overall, \textbf{\textsc{Idea-Catalyst}} provides a pathway for both AI-assisted human research and autonomous scientific discovery systems to engage in the kind of creative, boundary-spanning ideation process that drives breakthrough innovations. It addresses the critical gap in current automated scientific discovery methodologies by prioritizing the inherently exploratory, collaborative nature of research while providing systematic structure to cross-domain knowledge synthesis. Our main \textbf{contributions} are:
\begin{itemize}[leftmargin=5mm]
    \item We propose \textbf{\textsc{Idea-Catalyst}}, a metacognition-driven framework that systematically guides interdisciplinary scientific ideation through problem decomposition, cross-domain exploration, and strategic prioritization.
    \item We introduce a structured dataset and evaluation framework for benchmarking interdisciplinary research ideation across novelty, insightfulness, relevance, and usefulness.
    \item We show through LLM-based and human evaluations that \textsc{Idea-Catalyst} produces $21.38\%$ more novel and $16.22\%$ insightful ideas, while remaining grounded in the target research problem.
\end{itemize}


\begin{figure*}
    \centering
    \includegraphics[width=1.0\textwidth]{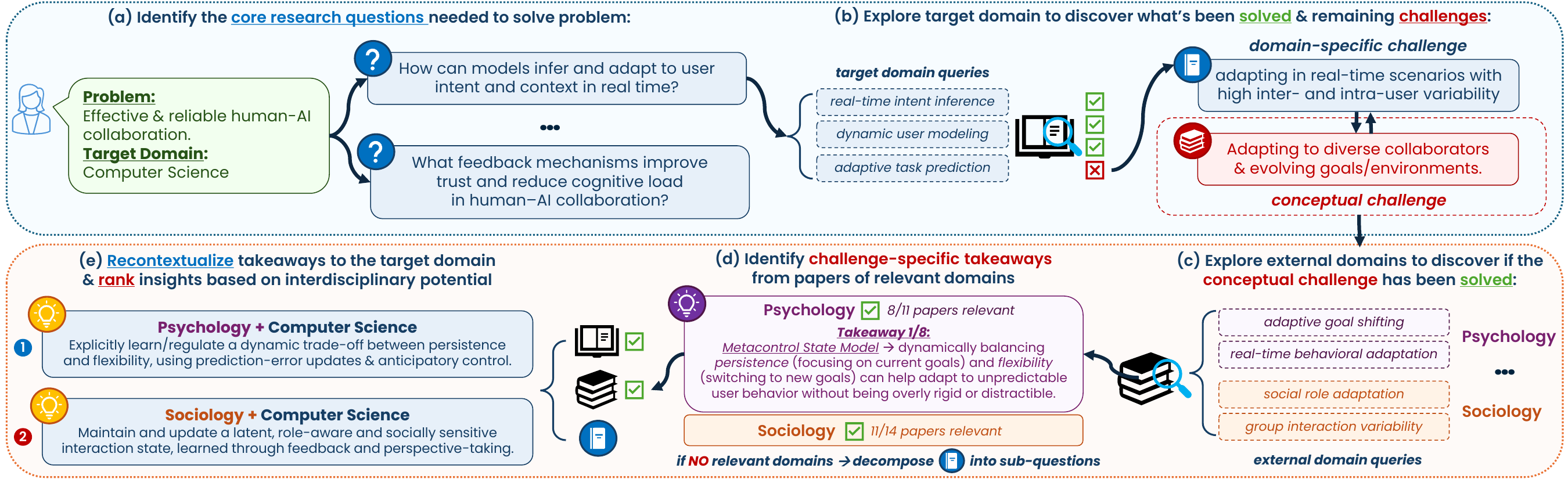}
    \caption{\textbf{\textsc{Idea-Catalyst}} is a metacognition-driven framework that \textit{(a)} analyzes target-domain progress, \textit{(b)} identifies unresolved challenges, \textit{(c)} explores source domains for analogous insights, and \textit{(d)} integrates them into interdisciplinary idea fragments. The figure is illustrated with a real case study (summarized).}
    \label{fig:framework}
    \vspace{-2mm}
\end{figure*}

\section{Related Work}

\paragraph{\textbf{Interdisciplinary Research and Scientific Creativity}}
Prior work in the science-of-science literature has shown that interdisciplinary research is a key driver of scientific innovation and long-term impact, with ideas that integrate concepts across distant fields often achieving substantially higher influence~\cite{van2015interdisciplinary,okamura2019interdisciplinarity}. Studies of creative processes further emphasize that scientific breakthroughs typically emerge through the gradual accumulation and recombination of partial ideas rather than isolated insights~\cite{sosa2019accretion,gonccalves2021life}, often drawing from multiple conceptual domains~\cite{yanai2019night}. Despite these benefits, deeply integrative interdisciplinary research remains rare and fragile~\cite{porter2009science,raasch2013rise}, in part because identifying which external domains are meaningfully relevant—and how their ideas can be translated into a target domain—poses a substantial cognitive and practical challenge for individual researchers.

\paragraph{\textbf{Automated Scientific Discovery and Research Ideation}}
Recent advances in large language models have spurred work on AI-assisted scientific discovery, including automated literature review, hypothesis generation, research ideation, and experimental planning~\cite{gottweis2025towards,goel2025training,si2026towards,kargupta-etal-2025-tree,kargupta-etal-2025-taxoadapt,jansen2025codescientist}. Systems such as SCIMON~\cite{wang2024scimon} and benchmarks like IdeaBench~\cite{guo2025ideabench} explore literature-grounded idea generation and evaluation, finding that while LLMs tend to produce highly novel ideas, they often lack technical depth, feasibility, or strong grounding in concrete research challenges~\cite{si2024can}. Moreover, many automated approaches tightly couple ideation with execution or early evaluation, which can bias exploration toward incremental, single-domain refinements and reduce cross-domain creativity~\cite{si2026towards}. In contrast, \textsc{Idea-Catalyst} explicitly avoids premature execution and instead focuses on supporting early-stage exploratory ideation through structured target-domain analysis and strategically guided interdisciplinary retrieval.

\paragraph{\textbf{Human-Centered Scientific Knowledge Discovery}}
Complementary to fully automated methods, human-centered systems leverage LLMs to support researchers in literature exploration, question formulation, and iterative refinement \cite{kargupta-etal-2025-beyond}. Tools such as IdeaSynth~\cite{pu2025ideasynth} and DiscipLink~\cite{zheng2024disciplink} demonstrate that interactive, human-in-the-loop approaches can effectively support exploratory thinking and interdisciplinary information seeking. However, these systems typically rely on either user input or the LLM’s parametric knowledge to suggest relevant domains, which can favor nearby or familiar fields and overlook deeper conceptual analogies across distant disciplines. \textsc{Idea-Catalyst} complements this line of work by introducing a metacognition-driven framework that explicitly decomposes target-domain problems, abstracts persistent conceptual challenges, and strategically guides cross-domain exploration---supporting not only information access, but the discovery of high-impact interdisciplinary insights.
\vspace{-2.5mm}

\section{Methodology}
\par \textbf{\textsc{Idea-Catalyst}} aims to augment early-stage scientific ideation by (a) decomposing research problems into core questions, (b) identifying unresolved conceptual challenges in the target domain, (c) extracting insights from external source domains which address these challenges, and (d) integrating them into an interdisciplinary idea fragment. The overall framework is illustrated in Figure~\ref{fig:framework}.

\subsection{Preliminaries}
\label{sec:preliminaries}

\subsubsection{\textbf{Problem Formulation}}

\par To support early-stage conceptual brainstorming, we assume as input only a short research problem statement $p$ (e.g., 1--2 sentences on effective and reliable human-AI collaboration) situated within a \emph{target domain} $\mathcal{D}_{\text{target}}$ (e.g., Natural Language Processing). Our \textit{objective} is to generate a set of interdisciplinary idea fragments $\mathcal{F}$, where each fragment $f_i \in \mathcal{F}$ is grounded in an external \emph{source domain} $\mathcal{D}_{s_i}$ and comprised of a set of literature-derived insights $t \in T_{s_i}$. Each fragment proposes a candidate interdisciplinary idea $\hat{T_{s_i}}$ by recontextualizing these insights to address $p$, thereby integrating concepts from $\mathcal{D}_{s_i}$ into the target domain  $\mathcal{D}_{\text{target}}$.

\begin{defn}[\textbf{\textit{Target Domain}}]
The \emph{target domain} $\mathcal{D}_{\text{target}}$ denotes the primary scientific field in which the research problem $p$ is situated. It is characterized by its established literature, methodologies, problem formulations, and evaluation norms. The goal of our framework is to augment ideation within $\mathcal{D}_{\text{target}}$ by introducing conceptually grounded insights originating outside the domain.
\label{def:target}
\end{defn}

\begin{defn}[\textbf{\textit{Source Domain}}]
A \emph{source domain} $\mathcal{D}_{s}$ is a scientific field distinct from the target domain  $\mathcal{D}_{\text{target}}$, characterized by its own literature, conceptual frameworks, and problem-solving traditions. Source domains serve as potential reservoirs of transferable insights that, when appropriately recontextualized, may help address unresolved conceptual challenges in $\mathcal{D}_{\text{target}}$. To promote non-trivial interdisciplinary connections, we restrict source domains to fields that are sufficiently distant from the target domain at a coarse-grained level of similarity (e.g., Computer Science and Psychology), and exclude closely related subfields (e.g., Natural Language Processing and Machine Learning).
\label{def:source}
\end{defn}

\begin{defn}[\textbf{\textit{Interdisciplinary Insight}}]
An \emph{interdisciplinary insight} $t \in T_{s_i}$ is a literature-grounded conceptual takeaway extracted from the source domain $\mathcal{D}_{s_i}$. Such insights typically describe mechanisms, principles, or abstractions that are not natively expressed in the target domain $\mathcal{D}_{\text{target}}$ but may become relevant when mapped onto the research problem $p$.    
\label{def:insight}
\end{defn}

\begin{defn}[\textbf{\textit{Interdisciplinary Potential}}]
\emph{Interdisciplinary potential} denotes the expected value of an idea fragment $f_i \in \mathcal{F}$ for advancing the research problem $p$ through cross-domain integration. It reflects a fragment’s ability to (i) address unresolved conceptual challenges in $\mathcal{D}_{\text{target}}$, (ii) introduce non-trivial perspectives from $\mathcal{D}_{s_i}$, and (iii) plausibly inspire novel research directions when recontextualized within the target domain. Idea fragments in $\mathcal{F}$ are ranked according to this potential.
\label{def:potential}
\end{defn}

\subsubsection{\textbf{Scientific Literature Snippet Retrieval}}
\label{sec:snippet_retrieval}

\par To ground interdisciplinary ideation in existing scientific knowledge, we retrieve literature snippets using the Semantic Scholar \emph{Snippets} API\footnote{\url{https://api.semanticscholar.org/api-docs/snippets}} \cite{Kinney2023TheSS}. Given a natural-language query and a coarse-grained scientific domain (Appendix~\ref{app:semantic_domains}), the Snippets API returns short, relevance-ranked text passages extracted from papers, along with their associated metadata. Given that Semantic Scholar performs the underlying document parsing, indexing, and query-snippet relevance matching internally, this allows us to treat snippet retrieval as a black-box operation and focus on structuring the research ideation process rather than optimizing retrieval models. For each query, we retrieve the top-$k$ papers within a specified domain and aggregate multiple snippets per paper when available. When snippet text is unavailable or degenerate (e.g., identical to the paper title), we fall back to retrieving the paper abstract to ensure minimal contextual grounding. Retrieved snippets are used as lightweight, fine-grained evidence for downstream analysis, enabling the identification of unresolved challenges and transferable conceptual insights across domains while maintaining scalability across diverse scientific fields.

\subsection{Metacognition-Driven Ideation}
\label{sec:metacognition_target}

\par Scientific research ideation is inherently creative, requiring ideas to be both novel and useful~\cite{yanai2019night}. A central component of this process is \emph{metacognition}: the ability to monitor, evaluate, and regulate one’s own reasoning during problem solving \cite{flavell1979metacognition,Yeung2012-pv,kitchner1983cognition}. Prior work shows that creative performance depends on the metacognitive awareness of which strategies are appropriate, when to apply them, and how to assess progress. Inaccurate metacognitive monitoring can disrupt how individuals guide and adjust their creative reasoning, whereas stronger metacognition is consistently associated with more effective creative problem solving~\cite{urban2025we}. Thus, we align our framework to the following metacognitive behaviors~\cite{kargupta2025cognitive}:

\begin{itemize}[leftmargin=5mm]
    \item \textbf{\textit{Self-awareness:}}  
    Assessing what is known, what remains uncertain, and which aspects of a research problem are both challenging and actionable. In our framework, this corresponds to evaluating how thoroughly different facets of problem $p$ have been addressed in the target-domain literature.

    \item \textbf{\textit{Context awareness:}}  
    Recognizing the assumptions, constraints, and norms that shape a problem. For our task, this includes recognizing target-domain limitations and identifying external source domains that may offer complementary perspectives.

    \item \textbf{\textit{Strategy selection:}}  
    Choosing reasoning strategies aligned with the nature of the problem. In practice, this involves selectively exploring disciplines that are well suited to particular challenges and open research questions of $p$ (e.g., \textit{control theory} for formalization, \textit{psychology} for learning behavior).

    \item \textbf{\textit{Goal management:}}  
    Maintaining and adapting intermediate objectives. This manifests as decomposing $p$ into research questions, prioritizing those with the greatest potential for conceptual advancement, and assessing progress made, post-ideation.

    \item \textbf{\textit{Evaluation:}}  
    Monitoring the quality \& promise of the reasoning process. Rather than prematurely enforcing feasibility, the ideation process should assess whether insights meaningfully address unresolved conceptual challenges.
\end{itemize}

\par Under this view, creative ideation emerges from the coordination of two complementary reasoning modes~\citep{johnson2010mental}: \textit{\textbf{critical reasoning}}, which emphasizes structured evaluation and analytical rigor, and \textit{\textbf{creative reasoning}}, which supports the generative synthesis of novel and valuable ideas~\citep{de2022reasoning,dwyer2025evaluation}. Our framework balances these modes by grounding ideation in a systematic analysis of the target domain while preserving space for exploratory, interdisciplinary reasoning that expands the solution space~\citep{wechsler2018creative,halpern2007nature}.

\subsection{Critical Reasoning over the Target Domain}

\par \textsc{Idea-Catalyst} initiates the creative ideation process with a systematic analysis of $\mathcal{D}_{\text{target}}$, grounding the model's self-awareness and context awareness in the current state of the literature (e.g., state-of-the-art approaches, technical/conceptual limitations). This analysis enables a structured, critical assessment of what has already been addressed, where progress is uneven, and which conceptual challenges remain unresolved. 

\par To steer ideation toward insights that are both novel and useful, we analyze $\mathcal{D}_{\text{target}}$ to identify aspects of $p$ that are weakly addressed and therefore offer the greatest potential for impact. We first decompose $p$ into a structured set of research questions $q_i \in \mathcal{Q}$, allowing us to examine the problem from multiple complementary perspectives that may exhibit varying levels of maturity in the existing literature. Each question $q_i$ is represented in two forms: a \textbf{\textit{domain-specific}} formulation $q_i^{\mathcal{D}}$, expressed in the language and assumptions of $\mathcal{D}_{\text{target}}$, and a corresponding \textbf{\textit{domain-agnostic}} formulation $q_i'$ that abstracts away academic jargon and implementation details. This dual representation enables precise assessment of progress within $\mathcal{D}_{\text{target}}$ while facilitating conceptual cross-domain comparison. For example, given the problem of ``\textit{effective and reliable human--AI collaboration}'' (Figure~\ref{fig:framework}), the resulting research questions include:

\begin{table}[h]
\centering
\footnotesize
\renewcommand{\arraystretch}{1.2}
\begin{tabular}{p{0.46\linewidth} p{0.46\linewidth}}
\toprule
\textbf{Domain-Specific Question ($q_i^{\mathcal{D}}$)} & \textbf{Domain-Agnostic Question ($q_1'$)} \\
\midrule
How can models be trained to dynamically infer and adapt to user intent and task context in real-time collaborative scenarios?
&
\textit{How can understanding of intent and context be updated through continuous interaction?} \\[2mm]
\hline
How should a system decide when to take initiative vs. defer to human to maintain well-calibrated autonomy \& control across contexts?
&
\textit{When should control be exercised versus withheld?} \\
\bottomrule
\end{tabular}
\caption{Output examples of domain-specific \& agnostic research question pairs for ``human-AI collaboration''.}
\label{tab:question_pairs}
\vspace{-5mm}
\end{table}

\par For each research question $q_i$, we generate a set of natural-language search queries that capture its domain-specific formulation $q_i^{\mathcal{D}}$ (e.g., \textit{real-time intent inference}, \textit{dynamic user modeling}). These queries are used to retrieve a set of relevant papers and associated literature snippets $\{d_1, \dots, d_k\} \subset \mathcal{D}_{\text{target}}$ (Section~\ref{sec:snippet_retrieval}). Based on the retrieved papers, we assess their relevance to $q_i$ and evaluate the extent to which the question has been addressed in the target domain (e.g., \emph{largely resolved}: $\mathcal{Q}_{resolved} \subseteq \mathcal{Q}$, \emph{partially addressed}: $\mathcal{Q}_{partial}$, or \emph{largely unexplored}: $\mathcal{Q}_{open}$). Crucially, this analysis surfaces remaining critical, non-incremental challenges $q_j^i$ that are explicitly stated or implicitly suggested by the literature and are not resolved by existing approaches. Each remaining challenge $q_j^i$ inherits the same dual representation as its parent question $q_i$, consisting of both a domain-specific and a domain-agnostic formulation, as shown in Table~\ref{tab:challenge_pairs}.


\begin{table}[h]
\centering
\footnotesize
\renewcommand{\arraystretch}{1.2}
\begin{tabular}{p{0.46\linewidth} p{0.46\linewidth}}
\toprule
\textbf{Domain-Specific Challenge ($q_{1}^{1,\mathcal{D}}$)} & \textbf{Domain-Agnostic Challenge ($q_{1}^{\prime 1}$)} \\
\midrule
How can a system adapt in real-time to high inter/intra-user variability?
&
\textit{How can behavior adapt to diverse collaborators \& evolving goals/environments?} \\
\bottomrule
\end{tabular}
\caption{Outputted dual representation of a remaining challenge $q_j^i$ in addressing $q_i$, derived from analysis of $\mathcal{D}_{\text{target}}$.}
\label{tab:challenge_pairs}
\vspace{-7mm}
\end{table}

\subsection{Creative Reasoning Across Source Domains}
\label{sec:source_domain_exploration}

\par Having identified weakly addressed research questions and unresolved conceptual challenges in the target domain $\mathcal{D}_{\text{target}}$, \textsc{Idea-Catalyst} transitions from critical reasoning to creative exploration. Rather than expanding the solution space indiscriminately, we use insights from the target-domain analysis to strategically guide cross-domain exploration toward directions with the greatest potential for non-incremental interdisciplinary insight. Specifically, we prioritize research questions $q_i \in \mathcal{Q}_{open}$ that remain largely unexplored in $\mathcal{D}_{\text{target}}$, as well as conceptual challenges $q^i_j \in \mathcal{Q}_{partial}$, where alternative perspectives have the most potential to contribute.

\par This exploration is driven by the domain-agnostic form of each selected question or challenge. By abstracting away target-domain terminology and implementation details, these formulations isolate the underlying conceptual gaps that remain unresolved (Table~\ref{tab:challenge_pairs}). Such abstractions are more likely to correspond to theoretical constructs, explanatory frameworks, or empirical phenomena studied in external fields, even when surface-level applications differ. Consequently, domain-agnostic questions form the basis for selecting candidate source domains and structuring cross-domain search.

\subsubsection{\textbf{Cross-Domain Retrieval}}
\par For each domain-agnostic question or challenge $q'$, we first identify a small set of external source domains that are plausibly relevant through analogy (e.g., how groups coordinate in Sociology vs.\ how teams coordinate in human--AI systems), shared mechanisms (e.g., adaptation through feedback in Psychology vs.\ Control Theory), or transferable principles (e.g., reasoning about uncertainty in Cognitive Science vs.\ machine learning), while explicitly excluding domains that are overly proximal to $\mathcal{D}_{\text{target}}$. This selection reflects the intuition that \textit{more distant} domains are likelier to contribute novel perspectives rather than incremental variations of existing approaches. For each selected source domain $\mathcal{D}_s$, we then generate a small set of search queries which reflect the domain-specific vocabulary of $\mathcal{D}_s$ (e.g., specific terminologies, frameworks $\rightarrow$ ``\textit{cognitive load theory}'' or ``\textit{social role adaptation}'' in Sociology).

\par Using these queries, we retrieve papers and their respective snippets (Section ~\ref{sec:snippet_retrieval}) from each source domain and analyze them to determine whether they provide meaningful conceptual insight into the challenge. For each source domain, we assess the relevance of retrieved papers, where we only extract literature-grounded conceptual takeaways $t_i \in T_{s_i}^{q'}$ from domains where the \textit{majority} of retrieved papers are relevant to conceptual question/challenge $q'$. This ensures that \textit{the overall insight $t_i$ of the source domain has sufficient grounding in its literature} and is not an isolated finding. Table \ref{tab:source_domain_takeaways} showcases a real example of an insight from Psychology on the ``\textit{human-AI collaboration} problem. Each insight is structured into a set of takeaways, where each contains the specific source domain concept and its underlying logic/perspective in understanding the question/challenge ${q'}_j^i$.

\begin{table}[h]
\centering
\footnotesize
\renewcommand{\arraystretch}{1.2}
\begin{tabular}{p{0.46\linewidth} p{0.46\linewidth}}
\toprule
\textbf{Source Domain Concept} & \textbf{How Does it Work?}\\
\midrule
\textbf{Metacontrol State Model}: Goal-directed behavior reflects a balance between persistence (maintaining current goals) and flexibility (switching goals when conditions change). &
Dynamic regulation between persistence and flexibility allows adaptive behavior that remains focused while responding efficiently to changing goals and environments \cite{castro2021fast}. \\
\hline
\textbf{Cognitive Control}: Control processes should be prospectively adjusted based on the expected frequency of goal switches. &
Anticipatory adjustment of cognitive control reduces the cost of goal switching, enabling smoother transitions and lower cognitive load under frequent change \cite{grahek2023cost}. \\
\hline
\textbf{Dynamic Goal Pursuit}: Goal pursuit can be supported through just-in-time adaptive interventions that monitor behavior and provide context-sensitive support. &
Real-time monitoring and adaptive feedback help sustain goal pursuit under variability by aligning support with evolving goals and situational demands \cite{o2024ecological}. \\
\bottomrule
\end{tabular}
\caption{Summarized conceptual takeaways $t_1, t_2 \in T_{\text{psychology}}$ for challenge ${q'}_1^{1}$ in Table \ref{tab:challenge_pairs} and their top identified paper.}
\label{tab:source_domain_takeaways}
\vspace{-5mm}
\end{table}

\par By grounding source domain exploration in domain-agnostic challenges and emphasizing conceptual relevance over methodological similarity, \textbf{\textsc{Idea-Catalyst}} enables the systematic discovery of interdisciplinary perspectives that meaningfully expand the ideation space. The resulting conceptual takeaways form the foundation for the subsequent recontextualization and integration stage, where insights from multiple source domains are mapped back into $\mathcal{D}_{\text{target}}$ to generate candidate interdisciplinary idea fragments.

\subsection{Target--Source Interdisciplinary Integration}
\label{sec:integration}

\par While interdisciplinary insights from source domains may already inspire researchers to explore new perspectives or refine existing ideas, we further examine whether such insights can be meaningfully integrated with the target domain $\mathcal{D}_{\text{target}}$. This integration step allows us to assess which insights extend beyond inspiration to support concrete, cross-domain synthesis, and thus exhibit strong interdisciplinary potential.

\paragraph{\textbf{Idea Fragments}} We define an \emph{idea fragment} as a structured, intermediate representation capturing a conceptual mechanism, principle, or strategy extracted from a source domain and recontextualized for the target domain. Formally, an idea fragment links:
(i) a target-domain research challenge with its relevant retrieved papers,
(ii) source-domain conceptual takeaways \& corresponding literature, and
(iii) a rationale describing how the takeaway could address the target-domain challenge.
Idea fragments are intentionally incomplete: they are designed to support creative ideation rather than prescribe full solutions, aiming to capture promising directions for synthesis by articulating how concepts from $\mathcal{D}_s$ could be combined with existing approaches in $\mathcal{D}_{\text{target}}$.

\paragraph{\textbf{Generating Integrated Fragments}.}
For each eligible question--source-domain pair $(q_i, \mathcal{D}_s)$, we integrate the most relevant source-domain takeaways and their corresponding papers with the relevant literature retrieved from $\mathcal{D}_{\text{target}}$. Integration is guided by three considerations: \textit{\textbf{(i)}} how target-domain methods and assumptions can be complemented by source-domain perspectives, \textit{\textbf{(ii)}} how the combined view addresses the specific challenge underlying $q_i$, and \textit{\textbf{(iii)}} how limitations of either domain are mitigated through synthesis. The outcome is an idea fragment $f_i$ that proposes a concrete pathway for interdisciplinary integration.

\paragraph{\textbf{Ranking Interdisciplinary Potential}.}
Because multiple idea fragments $\mathcal{F}=\{f_{i}\}$ may be generated for a given research problem $p$ (from different $q_i$ and $\mathcal{D}_s$, we introduce the notion of \emph{interdisciplinary potential} to prioritize fragments that are most likely to yield impactful cross-domain advances. Interdisciplinary potential reflects a fragment’s expected value along dimensions such as depth of integration, degree of multi-stage disciplinary engagement, innovation payoff, and balance between novelty and feasibility \cite{okamura2019interdisciplinarity, porter2009science}.

\par Rather than assigning absolute scores, we conduct pairwise comparisons between idea fragments. Given two fragments $f_a, f_b \in \mathcal{F}$, we assess which fragment exhibits stronger interdisciplinary potential based on the above criteria. Aggregating preferences across all pairwise comparisons yields a ranked ordering of $\mathcal{F}$, from strongest to weakest interdisciplinary potential. This relative evaluation avoids the need for a single scalar metric, while still prioritizing the most promising integrated ideas. By structuring interdisciplinary exploration through a metacognition-driven framework, \textsc{\textbf{Idea-Catalyst}} supports early-stage research ideation while preserving both novelty and rigor.

\section{Experimental Design}

\par We choose \texttt{Qwen3-14B} \citep{yang2025qwen3} as our primary model for experiments (\texttt{no-thinking} for efficiency, temperature = $0.7$), and gpt-oss-120b (temperature = $0.0$) \citep{agarwal2025gpt} for our LLM judge. We retrieve a maximum of 20 papers per round of retrieval and prune source domains where the majority ($50\%$) of papers are irrelevant to the research problem.

\subsection{Dataset}
\label{sec:datasets}

\par We evaluate \textsc{Idea-Catalyst} using the \textsc{CHIMERA} dataset~\cite{sternlicht2025chimeraknowledgebaseidea}, a collection of interdisciplinary research papers drawn from arXiv with annotated \emph{inspiration relations} between source and target domains. Each instance links a target-domain contribution (\texttt{target\_text}) to a distinct source-domain inspiration (\texttt{source\_text}), making it well suited for studying cross-domain knowledge transfer and interdisciplinary ideation. We select 400 instances where the source and target domains belong to different coarse-grained scientific fields, the annotated relation is \textit{inspiration}, both domains are explicitly specified, and the annotated problem context (which serves as our input $p$) does not leak the source insights. We further prevent knowledge leakage by restricting retrieval (Section~\ref{sec:snippet_retrieval}) to papers published strictly before the arXiv posting year of each instance.
\par To enable direct comparison between generated idea fragments and ground-truth interdisciplinary contributions, we pre-process each sample into a structured representation aligned with our framework’s output format (Appendix \ref{app:fragmentformat}) using \texttt{Qwen3-14B} (\texttt{no-thinking, temperature$=0$}). The model is strictly constrained to extract and reorganize information already present in the abstract and its annotated context, and ignoring any experimental results, thereby preserving the original interdisciplinary intent while enabling fair, structure-aligned evaluation. For the human study, we ask each participant to provide a brief (1--2 sentence) description of a research problem they are currently working on or have worked on previously, along with the corresponding target domain.

\begin{table*}[!ht]
\caption{Takeaway-level average win rates at top-$k$ ($k \in \{1,2,3\}$). Bold indicates best; $^\dagger$ denotes second-highest.}
\centering
\small
\renewcommand{\arraystretch}{1.1}
\setlength{\tabcolsep}{10pt}
\rowcolors{3}{rowgray}{white}
\begin{tabular}{lccccccccc}
\toprule
\multirow{2}{*}{\textbf{Method}} 
& \multicolumn{3}{c}{\textbf{Insightfulness}} 
& \multicolumn{3}{c}{\textbf{Relevance}} 
& \multicolumn{3}{c}{\textbf{Overall}} \\
\cmidrule(lr){2-4} \cmidrule(lr){5-7} \cmidrule(lr){8-10}
& @1 & @2 & @3 & @1 & @2 & @3 & @1 & @2 & @3 \\
\midrule
Free-Form Source
& 18.25 & 23.06 & 27.35
& 47.75 & 51.45 & 51.57
& 44.25 & 48.06 & 50.14 \\

Guided Dual
& 73.00 & 74.19 & 72.36
& \textbf{62.25} & 59.68 & 60.54
& \textbf{66.75} & 63.23 & 64.39 \\
\hline

\textbf{\textsc{Idea-Catalyst}} 
& \textbf{85.50} & \textbf{85.16} & 84.47$^\dagger$
& 60.25 & \textbf{62.42} & 61.25
& 63.75 & 66.45$^\dagger$ & 65.67 \\

$\times$ Decompose
& 84.00$^\dagger$ & 84.88$^\dagger$ & 83.80
& 60.75$^\dagger$ & 61.63 & \textbf{65.32}
& 65.25$^\dagger$ & \textbf{66.82} & \textbf{70.13} \\

$\times$ Potential Ranking
& 84.71 & 84.28 & \textbf{84.83}
& 59.40 & 59.97 & 61.13
& 64.16 & 64.18 & 65.03 \\

$+$ Conceptual Rewriting
& 82.00 & 83.71 & 83.19
& 59.75 & 61.94$^\dagger$ & 62.82$^\dagger$
& 66.25 & 67.10 & 66.95$^\dagger$ \\

\bottomrule
\end{tabular}
\label{tab:takeaway_eval}
\end{table*}

\begin{table*}[!ht]
\caption{Idea-level average win rates at top-$k$ ($k \in \{1,2,3\}$). Bold indicates best; $^\dagger$ denotes second-highest.}
\centering
\small
\renewcommand{\arraystretch}{1.1}
\setlength{\tabcolsep}{10pt}
\rowcolors{3}{rowgray}{white}
\begin{tabular}{lccccccccc}
\toprule
\multirow{2}{*}{\textbf{Method}} 
& \multicolumn{3}{c}{\textbf{Novelty}} 
& \multicolumn{3}{c}{\textbf{Usefulness}} 
& \multicolumn{3}{c}{\textbf{Overall}} \\
\cmidrule(lr){2-4} \cmidrule(lr){5-7} \cmidrule(lr){8-10}
& @1 & @2 & @3 & @1 & @2 & @3 & @1 & @2 & @3 \\
\midrule
Free-Form Source
& 13.50 & 17.26 & 19.80
& 35.50 & 39.84 & 40.31
& 30.50 & 34.52 & 35.19 \\

Guided Dual
& 68.00 & 70.32 & 67.95
& \textbf{70.25} & 65.48$^\dagger$ & 64.39
& \textbf{72.25} & 68.39 & 66.10 \\
\hline

\textbf{\textsc{Idea-Catalyst}} 
& \textbf{83.25} & \textbf{84.03} & \textbf{83.05}
& 65.75$^\dagger$ & \textbf{66.13} & \textbf{66.38}
& 71.25$^\dagger$ & \textbf{70.65} & \textbf{70.09} \\

$\times$ Decompose
& 82.00$^\dagger$ & 81.94 & 78.48
& 61.00 & 63.43 & 65.06$^\dagger$
& 63.75 & 65.91 & 67.85$^\dagger$ \\

$\times$ Potential Ranking
& 83.21$^\dagger$ & 83.31$^\dagger$ & 81.79$^\dagger$
& 62.41 & 65.15$^\dagger$ & 64.16
& 68.17 & 69.04$^\dagger$ & 67.92 \\

$+$ Conceptual Rewriting
& 82.00 & 82.90 & 80.91
& 62.75 & 62.90 & 61.54
& 66.75 & 66.29 & 64.96 \\

\bottomrule
\end{tabular}
\label{tab:idea_eval}
\end{table*}

\subsection{Baselines}
\label{app:baselines}

\begin{table*}[!h]
\centering
\footnotesize
\caption{\textbf{Qualitative Comparison of Source-Domain Takeaways.} Exact outputs taken from \textit{Guided Dual} and \textsc{Idea-Catalyst}.}
\label{tab:case_study_exact}
\renewcommand{\arraystretch}{1.15}
\setlength{\tabcolsep}{5pt}
\begin{tabular}{p{1.5cm} p{5.25cm} p{9.9cm}}
\toprule
\textbf{Method}
& \textbf{Source Domain Framing} 
& \textbf{Applicability to Target Domain} \\
\midrule

\textbf{Guided Dual}
& Theory of Mind (ToM) and its role in predicting others' mental states 
& By equipping AI with ToM capabilities, the system can better understand and predict user intentions, leading to more natural and effective collaboration. This reduces cognitive load by aligning AI behaviors with user expectations and improving task efficiency. \\

\midrule
\textbf{\textsc{Idea-Catalyst}}
& Reciprocal information flow and role distribution enhance joint action coordination by allowing individuals to dynamically assign and shift roles based on task demands and the predictability of others' actions. 
& When individuals engage in reciprocal information flow, they can dynamically assign roles (e.g., leader--follower) and adjust their strategies in real time based on the actions and predictability of their partner. This enables them to adapt to complex and undefined tasks without predefined boundaries. \\

\bottomrule
\end{tabular}
\end{table*}

We compare \textsc{Idea-Catalyst} against two baselines that reflect common LLM-driven approaches to ideation with increasing degrees of retrieval structure: (a) \textit{\textbf{Free-Form Source Retrieval}} \cite{zheng2024disciplink} prompts the model to directly identify source domains without constraints on distance to the target domain, retrieve related literature, and synthesize ideas, without explicit target-domain analysis or problem decomposition; (b) \textit{\textbf{Guided Dual-Retrieval}} first retrieves representative target-domain literature and then conditions cross-domain retrieval (with explicit distance constraints) and ideation on this context, but does not explicitly identify unresolved conceptual challenges, construct domain-agnostic abstractions, or strategically guide source-domain selection. Both effectively serve as retrieve-then-ideate pipelines without metacognitive control.

\paragraph{\textbf{Ablations}} We evaluate three ablations to assess the contribution of individual components of \textsc{Idea-Catalyst}: \textit{$\times$ Decomposition}, which removes explicit problem decomposition and target-domain retrieval; \textit{$\times$ Interdisciplinary Ranking}, which replaces the potential-based ranking with the proportion of relevant source papers; and \textit{$+$ Conceptual Rewriting}, which rewrites final outputs for further clarity without altering technical content or structure. Together, these ablations isolate the effects of structured target-domain analysis, comparative interdisciplinary potential-based ranking, and articulation quality. We include further details in Appendix \ref{app:baselines}.

\subsection{Evaluation Metrics}

Although evaluation for creative ideation in scientific discovery is inherently subjective and requires high levels of domain expertise \citep{si2025ideation}, LLMs have shown strong capabilities in evaluating along various dimensions, while aligning with human judgements and capturing nuanced feedback signals \citep{lee2023rlaif, madaan2023self, mehri2025beyond, mehri2023automatic}. In particular, LLM judges have proven effective in evaluating creativity, demonstrating alignment with human judgments across diverse creativity dimensions and research ideation tasks \citep{si2024can, hou2025creativityprism, zhao2025assessing}.

We follow established practices in preference-based evaluation \citep{dubois2024length, zheng2023judging} and conduct a comparative evaluation against the ground truth. For each sample, we present both the generated output and the ground truth to an LLM judge, which determines which output better satisfies the evaluation criteria. We report the overall win rate against the ground truth across all samples, where a win indicates that the generated output is preferred over the ground truth.

We evaluate each sample along two dimensions:

\paragraph{\textbf{(1) Takeaways.}} We assess the quality of source-domain insights extracted by our approach. Specifically, we evaluate takeaways based on (1) interdisciplinary insightfulness, which measures whether the takeaways introduce specific, non-obvious concepts or frameworks from the source domain that are intellectually interesting to researchers in the target domain, and (2) interdisciplinary relevance, which assesses whether the takeaways have strong potential to inspire new approaches or address gaps in the target domain. 

\paragraph{\textbf{(2) Idea.}} We assess the quality of the final generated idea that integrates interdisciplinary source-domain insights for the target domain. Each idea is evaluated based on (1) interdisciplinary novelty, which measures whether the novelty of the idea, and (2) interdisciplinary usefulness, which measures which idea has greater potential for addressing the research problem in the target domain.

The complete evaluation prompts are provided in Appendix \ref{app:eval_prompts}. This evaluation approach allows us to assess the practical utility of \textsc{Idea-Catalyst} for supporting early-stage research ideation and provides a methodology that can be adopted for evaluating creativity in idea generation in future work.

\section{Experimental Results}

\paragraph{\textbf{Overall Performance \& Analysis}.}
Across both evaluation levels, \textsc{Idea-Catalyst} consistently outperforms baseline approaches on dimensions associated with exploratory and creative ideation. At the takeaway level (Table~\ref{tab:takeaway_eval}), it achieves the highest insightfulness scores across all top-$k$ settings, corresponding to an average relative improvement of \textbf{16.22\%} over \textit{Guided Dual} and \textbf{282.21\%} over \textit{Free-Form Source} when averaged across $k \in \{1,2,3\}$. Similarly, at the idea level (Table~\ref{tab:idea_eval}), \textsc{Idea-Catalyst} attains the strongest novelty scores, yielding an average relative gain of \textbf{21.38\%} over \textit{Guided Dual} and \textbf{407.65\%} over \textit{Free-Form Source}. This suggests that metacognition-driven ideation--strategically guided creative exploration grounded in target-domain critical reasoning--enables the model to surface ideas and takeaways that are more novel and insightful than purely retrieval-based baselines. We also observe a trade-off between exploratory creativity and grounded problem applicability: the interdisciplinary potential-based ranking generally favors novelty and insightfulness, with relevance and usefulness increasing as $k$ grows. When these criteria conflict, the overall LLM-judged score tends to prioritize relevance and usefulness, explaining why baselines can remain competitive on overall metrics despite substantially lower novelty or insightfulness.

\paragraph{\textbf{Ablation Studies}.}
Ablation results further highlight the contribution of each component in \textsc{Idea-Catalyst}. Removing target-domain decomposition leads to reduced novelty and insightfulness, likely because the model lacks a structured understanding of unresolved conceptual gaps and defaults to more surface-level ideas. Replacing interdisciplinary potential-based ranking with relevance-based heuristics similarly lowers overall performance, emphasizing the importance of pairwise comparative evaluation for identifying high-impact interdisciplinary ideas as opposed to an isolated relevance metric. Finally, while \textit{Conceptual Rewriting} does not yield quantitative gains, it improves clarity and interpretability; qualitative analysis and human studies indicate that outputs are often verbose and struggle to concisely communicate cross-domain insights while preserving essential technical content.

\begin{figure}[h]
    \centering
    \includegraphics[width=1.0\linewidth]{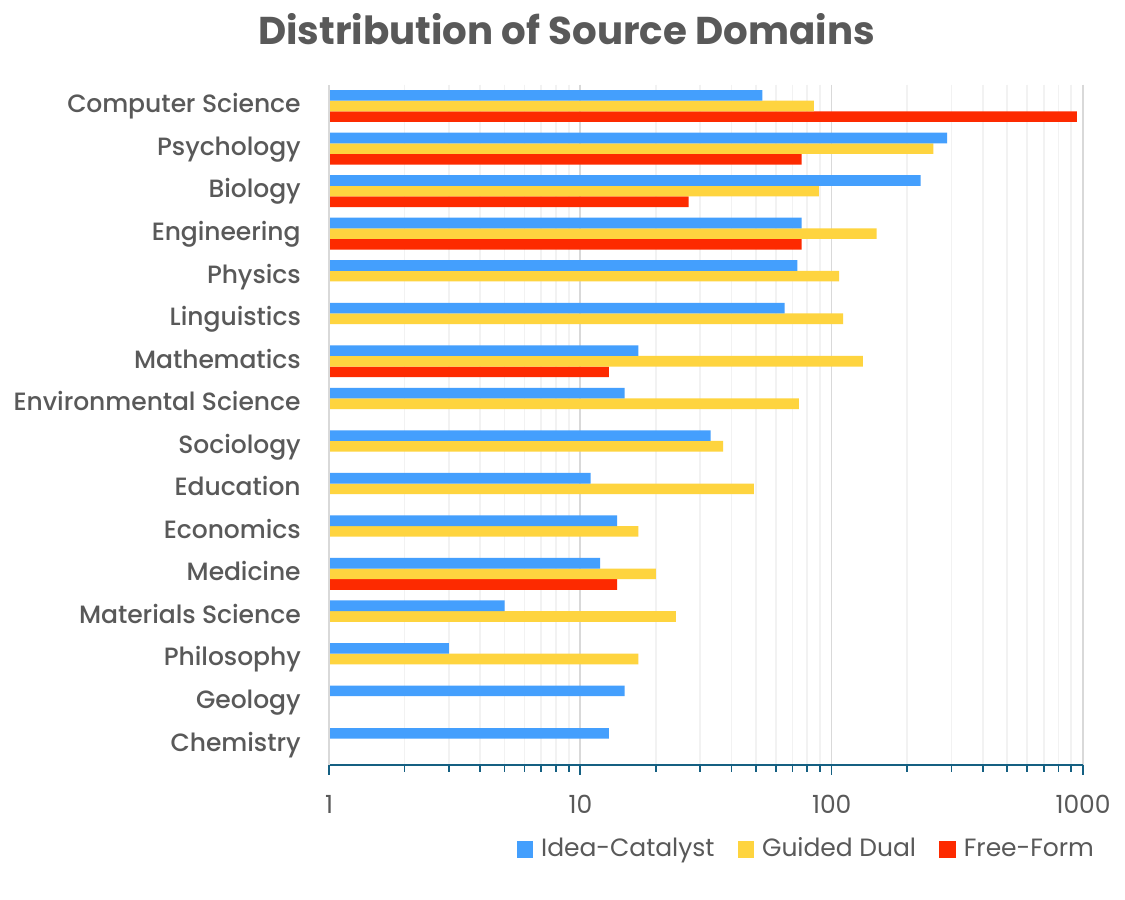}
    \caption{Source-domain distributions (log-scale) for each method's top three ideas.}
    \label{fig:source_dist}
    \vspace{-5mm}
\end{figure}

\begin{figure}[h]
    \centering
    \includegraphics[width=0.85\linewidth]{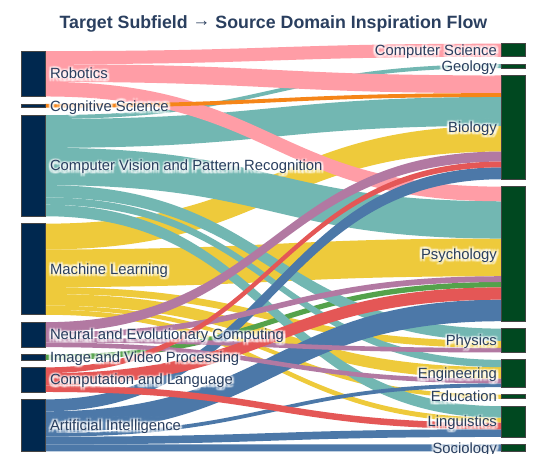}
    \caption{Target-source flow of interdisciplinary inspiration.}
    \label{fig:sankey}
    \vspace{-5mm}
\end{figure}

\subsection{Source Domain Distribution Analysis}

\par We analyze the distribution of source domains selected by each method by aggregating the top-$3$ ideas per problem and filtering out domains with fewer than $10$ occurrences (Figure~\ref{fig:source_dist}). The \textit{Free-Form Source Retrieval} baseline exhibits a severe skew toward \textit{Computer Science} (947 occurrences), resulting in very low domain diversity (normalized entropy $H_{\text{norm}} = 0.326$), indicating that unconstrained LLM-driven ideation tends to remain within the target domain’s immediate conceptual neighborhood even when encouraged to explore externally. In contrast, \textit{Guided Dual-Retrieval} achieves the highest overall spread across domains such as Psychology, Engineering, Mathematics, Linguistics, and Physics ($H_{\text{norm}} = 0.812$), although it also favors closer domains, such as Computer Science and Engineering ($19.67\%$ of its source domains vs. \textsc{Idea-Catalyst}'s $10.75\%$). \textsc{Idea-Catalyst} exhibits broad cross-domain exploration spanning Psychology, Biology, Physics, Linguistics, Engineering, and additional scientific fields ($H_{\text{norm}} = 0.682$), while consistently achieving competitive or higher relevance/usefulness alongside substantially stronger novelty/insightfulness (Table~\ref{tab:takeaway_eval} \&~\ref{tab:idea_eval}). Overall, this suggests that effective interdisciplinary ideation depends not only on increasing domain diversity, but on critically selecting source domains which yield meaningful conceptual insights.

\paragraph{\textbf{Target--Source Flow of Inspiration}} 
Figure~\ref{fig:sankey} visualizes the flow of interdisciplinary inspiration between target subfields and source domains, restricted to target--source pairs occurring at least 10 times (top-10 sources per target) to highlight stable patterns. \textit{Psychology} emerges as the most prevalent source across many AI-related targets, reflecting its foundational role in cognition, decision-making, and human--AI interaction. We also observe intuitive alignments, such as \textit{Neural and Evolutionary Computing} drawing from \textit{Biology}, and \textit{Artificial Intelligence} sourcing from both \textit{Psychology} and \textit{Linguistics}. Overall, the diagram shows that \textsc{Idea-Catalyst} surfaces diverse yet intuitive cross-domain influences, with different Computer Science subfields drawing from complementary external disciplines rather than a single dominant source.

\subsection{Qualitative Comparison of Takeaways}

\par Table~\ref{tab:case_study_exact} presents a qualitative comparison of the top-ranked source-domain takeaways selected by \textit{Guided Dual} and \textsc{Idea-Catalyst} for the problem of human--AI collaboration in open-ended tasks. While both methods identify \textit{Psychology} as the most relevant source domain, the nature of the extracted takeaways differs substantially. \textit{Guided Dual} selects a broadly applicable \emph{Theory of Mind} formulation that reflects well-established, high-level psychological concepts, but remains relatively generic and loosely tied to the specific challenges of open-ended, co-creative collaboration. In contrast, \textsc{Idea-Catalyst} surfaces a more targeted and problem-aligned takeaway centered on \emph{reciprocal information flow} and \emph{dynamic role distribution}, directly addressing coordination, role adaptation, and learning dynamics central to the research problem. This contrast suggests that beyond identifying relevant source domains, \textsc{Idea-Catalyst}’s metacognitive guidance enables more precise extraction of interdisciplinary insights meaningful for the target domain.

\subsection{Human Study}
\label{sec:human_study}

\par We conducted a human study with six PhD researchers working in Machine Learning, Natural Language Processing, and Electrical Engineering, each of whom provided a real research problem drawn from their own work (participant details and evaluation interface details shown in Appendix~\ref{app:human_study}). Overall, participants found \textsc{Idea-Catalyst} to be a useful ideation aid, particularly in identifying meaningful research questions and surfacing interdisciplinary perspectives. On average, researchers rated the relevance of the generated research questions highly (4.00/5), indicating that the system effectively captured core challenges in their problem formulations. Retrieved papers were also rated favorably (3.50/5), suggesting that retrieval was generally aligned with the researchers’ interests and problem context.

At the level of source-domain reasoning, takeaways were rated as moderately relevant (3.13/5) and insightful (3.16/5), with especially positive feedback from researchers working on problems that are naturally interdisciplinary (e.g., persuasion susceptibility of LLMs). Several participants reported that the takeaways introduced concepts they were motivated to further explore independently. Interpretability received an intermediate score (2.78/5), indicating that rationales were generally clear or mostly clear, with only minor ambiguity on average. Despite this, post-study interviews revealed that participants still perceived the takeaways and ideas as verbose, even after one round of conceptual rewriting. This highlights a central challenge in interdisciplinary ideation: balancing accessibility and brevity with the need to preserve critical technical and conceptual detail when translating ideas across domains. Future work could explore personalization strategies that adapt the level of abstraction and explanation to a user’s background and target domain.

Finally, participants rated generated ideas as more novel (3.22/5) than useful (3.00/5), echoing prior findings in \cite{si2025ideation}, namely that creative research ideas do not always translate directly into immediately actionable solutions. Taken together, these results suggest that \textsc{Idea-Catalyst} is effective at supporting early-stage exploratory thinking and conceptual reframing, while also revealing opportunities to further improve conciseness, grounding, and user-adaptive explanation in future iterations.

\section{Conclusion}

In this work, we introduced \textsc{Idea-Catalyst}, a metacognition-driven framework for interdisciplinary research ideation that supports targeted, creative exploration across scientific domains. By explicitly analyzing target-domain challenges and strategically guiding cross-domain inspiration, \textsc{Idea-Catalyst} surfaces ideas and insights that are significantly more novel and insightful than existing baselines, while remaining grounded in the original research problem. Our results suggest that supporting the \emph{process} of interdisciplinary ideation is a promising direction for human–AI collaboration in scientific research. Future work includes developing personalized summarization strategies tailored to researchers’ backgrounds, as well as leveraging interdisciplinary signals to recommend potential collaborators across domains.



\bibliographystyle{ACM-Reference-Format}
\bibliography{sample-base}

\appendix

\section{Idea Fragment Output Format}
\label{app:fragmentformat}

We represent each idea as an \emph{idea fragment} with the following schema:

\begin{tcolorbox}[
    colback=gray!5!white,
    colframe=gray!75!black,
    title={\textbf{Idea Fragment Format}},
    left=2mm,
    right=2mm,
    top=2mm,
    bottom=2mm,
    breakable,
    enhanced
]

\begin{Verbatim}[fontsize=\footnotesize]
  "idea_fragment": {
    "title": "Brief, descriptive title (max 15 words)",
    "core_insight": "2–3 sentence summary of the integration",
    "integration_mechanism": {
      "target_domain_elements": [
        "Target-domain concept or method",
        "Another target-domain concept or method"
      ],
      "selected_takeaways": [
        {
          "takeaway_id": "t1",
          "source_domain_formulation":
            "Conceptual insight using source-domain framing",
          "mechanism_explanation":
            "Explanation of the underlying conceptual logic",
          "selection_rationale":
            "Why this takeaway is relevant for integration"
        }
      ],
      "synthesis_approach":
        "Description of how elements are combined"
    },
    "challenge_resolution": {
      "addresses_target_challenge":
        "How the integration addresses the challenge",
      "addresses_source_limitations":
        "How integration mitigates limitations of the insight",
      "addresses_research_problem":
        "How this contributes to the overall research problem"
    },
    "concrete_realization": {
      "proposed_approach":
        "Specific algorithm, or technical realization",
      "key_innovations": [
        "Novel aspect enabled by integration",
        "Additional emergent innovation"
      ]
    }
  }
}
\end{Verbatim}
\end{tcolorbox}

\section{Baselines}
\label{app:baselines}
We compare \textsc{Idea-Catalyst} against two baseline methods that reflect common LLM-driven approaches to interdisciplinary ideation with increasing degrees of retrieval structure:
\begin{itemize}
    \item \textit{\textbf{Free-Form Source Retrieval}} \cite{zheng2024disciplink} prompts the model to directly identify potentially relevant source domains (with no restriction on distance to the target domain) for a given research problem, generate search queries, retrieve papers from those domains, and synthesize research ideas, without explicit analysis of the target domain or decomposition of the problem. This baseline captures an intuition-driven approach that relies primarily on the model’s parametric knowledge of the target domain rather than systematic reasoning about research gaps. 
    \item \textit{\textbf{Guided Dual-Retrieval}} introduces additional structure by first retrieving representative literature from the target domain, then conditioning cross-domain exploration and ideation on this retrieved context. While this baseline incorporates retrieval from both target and source domains, it does not explicitly identify unresolved conceptual challenges, construct domain-agnostic abstractions, or strategically guide source-domain selection, effectively serving as a retrieve-then-ideate pipeline without metacognitive control.
\end{itemize}

\begin{table*}[t]
\centering
\small
\caption{Participant backgrounds and research problems used in the human study. All participants are PhD researchers ranging from 3 to 5 years of research experience.}
\label{tab:human_backgrounds}
\renewcommand{\arraystretch}{1.15}
\setlength{\tabcolsep}{8pt}
\begin{tabular}{p{3.5cm} p{4cm} p{8.5cm}}
\toprule
\textbf{Primary Research Area} & \textbf{Target Domain} & \textbf{Research Problem Description} \\
\midrule

Natural Language Processing 
& Multilingual NLP 
& Are there tasks where LLM performance varies systematically across languages, particularly for culture-specific queries, and how can such performance disparities be mitigated? \\

Electrical Engineering 
& In-Memory Computing 
& How can the accuracy of in-memory computing systems be improved while preserving high energy efficiency, especially for Edge-AI applications? \\

Natural Language Processing 
& Multilingual Semantics 
& Why do language models change their answers across languages for the same query, even in high-resource settings, and does this reflect knowledge or semantic misalignment? \\

Natural Language Processing 
& Persuasion and Safety 
& How can we characterize and mitigate the susceptibility of LLMs to persuasion, including harmful or adversarial influence, while preserving beneficial adaptability? \\

Machine Learning
& Model Interpretability 
& How can influence functions be made dynamic, such that the importance of data points adapts across models and training contexts rather than remaining static? \\

Natural Language Processing 
& User Simulation 
& How can LLM-based user simulators better maintain consistent personas, reflect diverse user behaviors, and be evaluated for realism in multi-turn interactions? \\

\bottomrule
\end{tabular}
\end{table*}

\paragraph{\textbf{Ablations}} We conduct the following ablation studies to isolate the contributions of key components of \textsc{Idea-Catalyst}:
\begin{itemize}
    \item \textit{\textbf{No Decomposition}} removes the target-domain decomposition stage, relying instead on the model’s parametric knowledge to assess which aspects of the research problem have been addressed and which challenges remain, without explicitly decomposing the problem into research questions or retrieving target-domain literature conditioned on them. This evaluates the importance of structured, retrieval-grounded self- and context-awareness for identifying meaningful research gaps.
    
    \item \textit{\textbf{No Interdisciplinary Ranking}} removes the interdisciplinary potential-based ranking stage, replacing pairwise LLM-based comparisons with a heuristic ranking based solely on the proportion of retrieved source-domain papers deemed relevant to the target challenge. This tests whether explicit comparative evaluation is necessary to surface high-impact interdisciplinary ideas beyond relevance alone.
    
    \item \textit{\textbf{Conceptual Rewriting}} retains the full pipeline but replaces the final idea fragment with a rewritten version that improves conceptual clarity and accessibility while preserving structure, technical content, and domain grounding. This assesses whether observed gains stem from deeper interdisciplinary integration rather than improved articulation or presentation.
\end{itemize}

\section{Domains Supported by Semantic Scholar for Retrieval}
\label{app:semantic_domains}

\par Our literature retrieval pipeline relies on the Semantic Scholar \emph{Snippets} API, which supports search queries over a fixed set of coarse-grained scientific domains. Specifically, Semantic Scholar indexes papers under the following fields of study: \textit{Computer Science, Medicine, Chemistry, Biology, Materials Science, Physics, Geology, Psychology, Art, History, Geography, Sociology, Business, Political Science, Economics, Philosophy, Mathematics, Engineering, Environmental Science, Agricultural and Food Sciences, Education, Law}, and \textit{Linguistics}.

\par To ensure compatibility with this constraint, all fine-grained target-domain subfields provided as input to \textsc{Idea-Catalyst} (e.g., \textit{Natural Language Processing}, \textit{Reinforcement Learning}, \textit{Cognitive Science}) are mapped to their corresponding coarse-grained Semantic Scholar domains (e.g., \textit{Computer Science}, \textit{Psychology}) prior to retrieval. This mapping is used \emph{solely for literature retrieval and filtering} and does not affect the conceptual formulation of research questions, domain-agnostic abstractions, or subsequent interdisciplinary reasoning stages.

\section{Human Study Details \& Participant Backgrounds}
\label{app:human_study}

\par
\begin{figure*}
    \centering
    \includegraphics[width=1.0\textwidth]{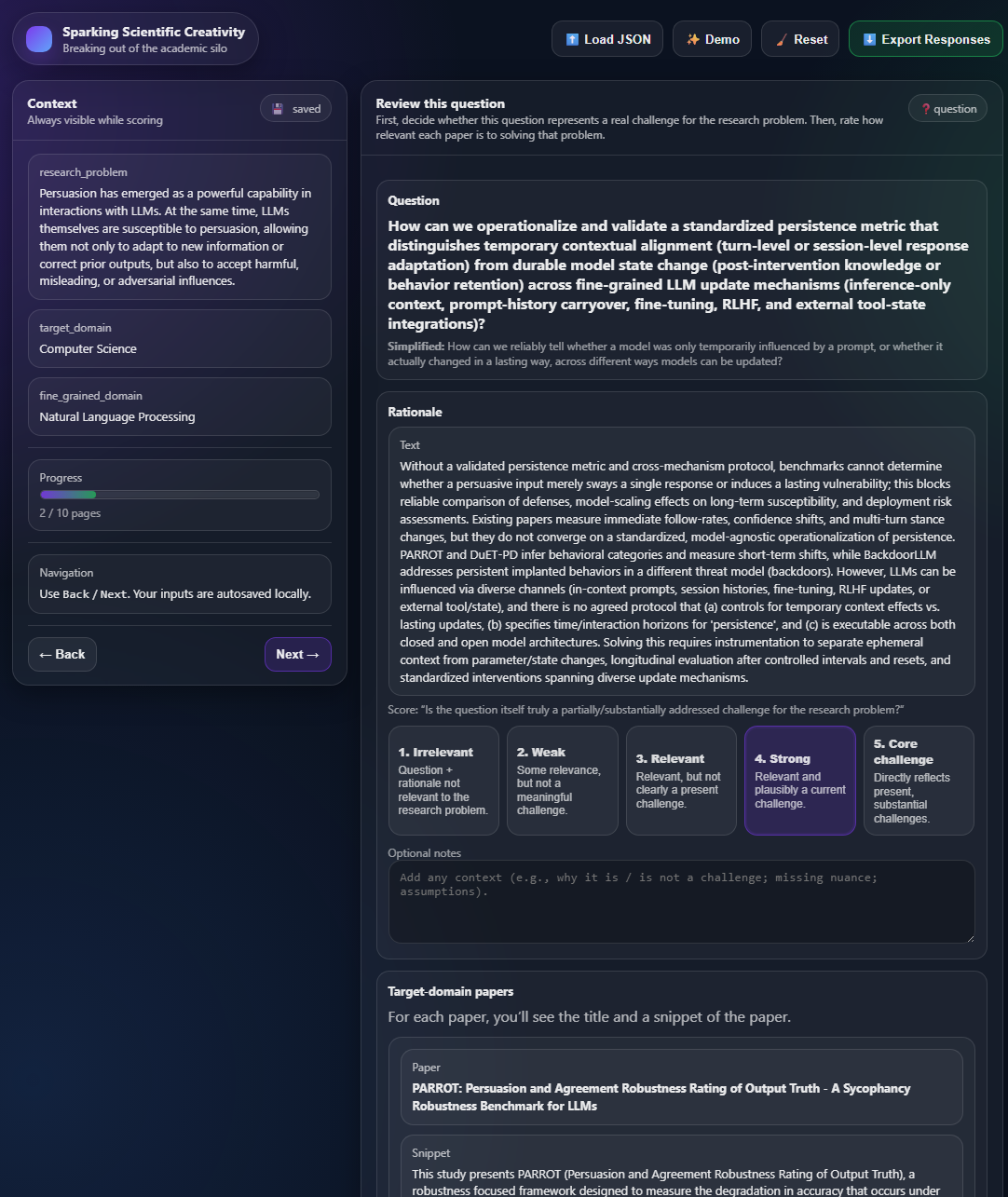}
    \caption{Screenshot of evaluation interface (reviewing questions/challenges).}
    \label{fig:interface_1}
\end{figure*}

\begin{figure*}
    \centering
    \includegraphics[width=1.0\textwidth]{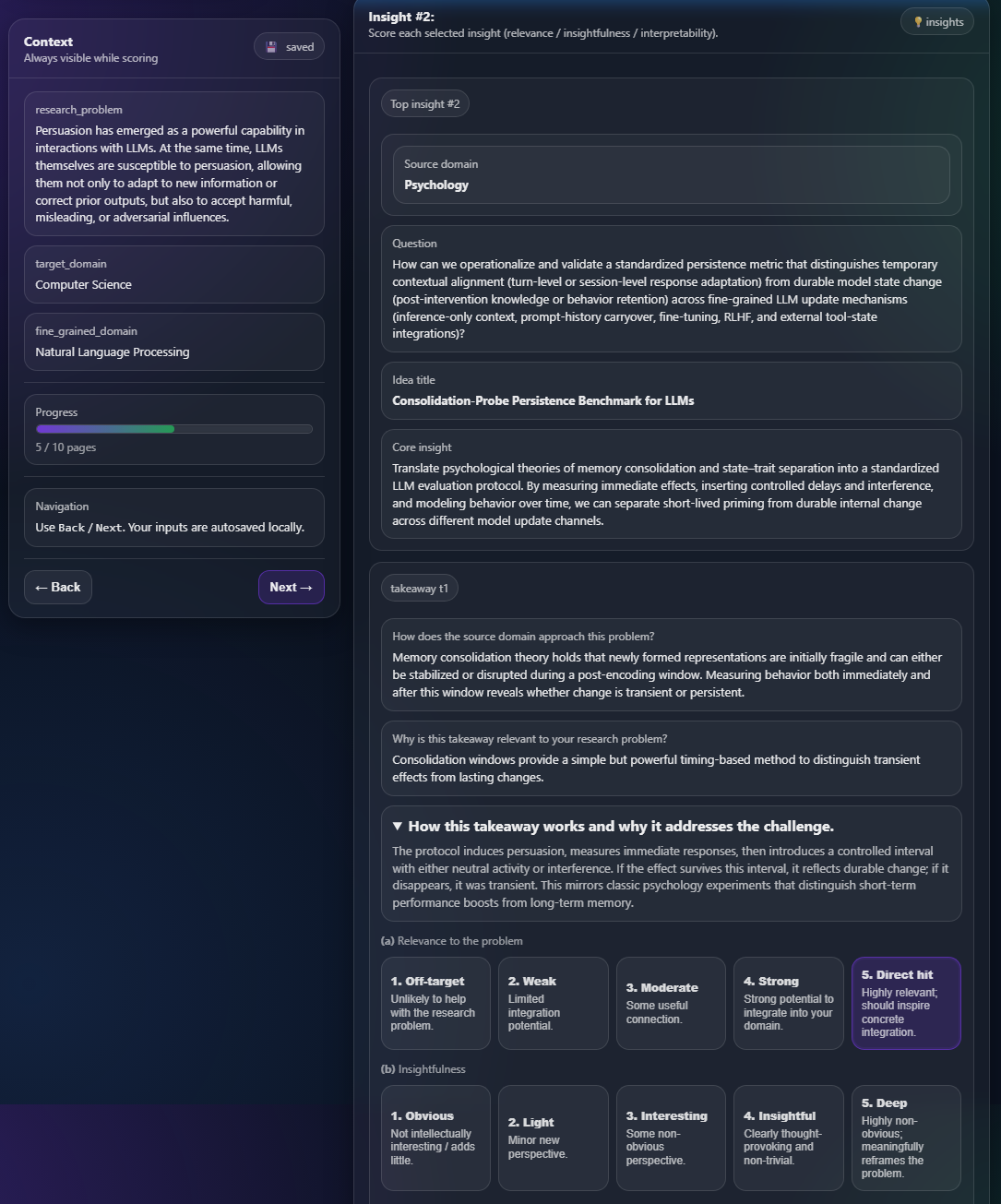}
    \caption{Screenshot of evaluation interface (reviewing takeaways).}
    \label{fig:interface_2}
\end{figure*}

\begin{figure*}
    \centering
    \includegraphics[width=1.0\textwidth]{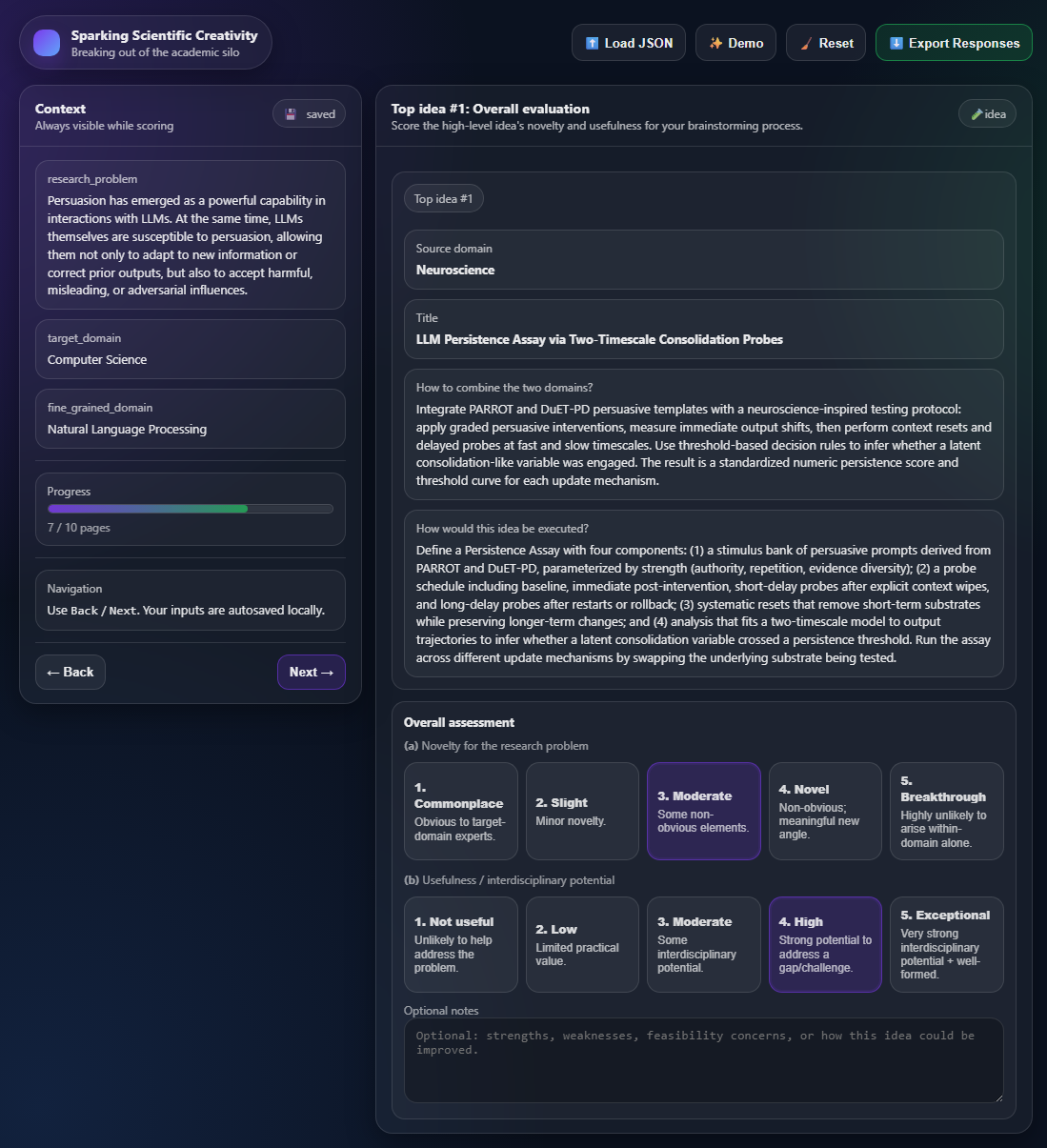}
    \caption{Screenshot of evaluation interface (reviewing ideas).}
    \label{fig:interface_3}
\end{figure*}

 We conducted a human study with six PhD researchers working in Machine Learning, Natural Language Processing, and Electrical Engineering, each of whom provided a real research problem drawn from their own work (as shown in Table \ref{tab:human_backgrounds}). We note that the study was declared exempt after being reviewed by our Institutional Review Board (IRB). We provide screenshots of our evaluation interface in Figures \ref{fig:interface_1}, \ref{fig:interface_2}, and \ref{fig:interface_3}.

\section{Evaluation Prompts}
\label{app:eval_prompts}

\begin{tcolorbox}[
    colback=gray!5!white,
    colframe=gray!75!black,
    title={\textbf{Takeaway Evaluation Prompt}},
    left=2mm,
    right=2mm,
    top=2mm,
    bottom=2mm,
    breakable,
    enhanced
]

\begin{Verbatim}[fontsize=\footnotesize, breaklines=true,breakanywhere=true]
You are an expert evaluator assessing the quality of cross-domain research takeaways.
Your task is to compare takeaways from two different methods that attempt to address the 
same research problem by drawing insights from domains outside the target domain.

--------------------------------------------------
RESEARCH PROBLEM
{research_problem}

TARGET DOMAIN
{target_domain}

--------------------------------------------------
METHOD 1 TAKEAWAYS
{method_1_text}

--------------------------------------------------
METHOD 2 TAKEAWAYS
{method_2_text}

--------------------------------------------------
EVALUATION CRITERIA

When evaluating Method 1 and Method 2, explicitly ground your judgment in the relevant 
fields of each takeaway, as described below.

### 1. INTERDISCIPLINARY INSIGHTFULNESS
Assess whether the method's takeaways provide insightful perspectives on the research 
problem.
  - Perspectives should introduce specific concepts/frameworks from their respective 
    source domain
  - Insightful perspectives should be intellectually interesting, non-   obvious, and 
    thought-provoking to researchers in the target domain ({target_domain})
    - Non-obvious perspectives typically come from source domains that are meaningfully 
      distinct from the target domain ({target_domain})

### 2. INTERDISCIPLINARY RELEVANCE
Assess whether the method's takeaways are relevant to the research problem and have 
strong potential for integration in the target domain ({target_domain}).
  - Ideal takeaways should:
    - Inspire new approaches/solutions to the research problem in the target domain 
    ({target_domain})
    - Address a gap/challenge for the research problem in the target domain 
    ({target_domain})
  - The complexity, simplicity, or practicality of the takeaway should not factor into 
  your decision (e.g., a "clear, immediately applicable" solution does not mean more 
  relevant). Relevance is defined based on the potential impact of the source domain 
  being introduced to the target domain for the research problem.
  - Keep in mind that if the distance between the source and target domain is larger 
  (e.g., Computer Science & Engineering are closer than Computer Science & Philosophy), 
  the idea may inherently be less practical. This does not mean that it is less relevant. 
  Focus on the degree of the potential impact to the research problem instead.

IGNORE:
- Length of explanations
- Narrative polish
- Missing implementation details

CONSIDER:
- **Consistency**: Are the method’s takeaways consistently 
meaningful, or uneven?
- **Groundedness**: Are claims supported by real conceptual 
alignment?
- **Scope appropriateness**: Are takeaways neither trivial 
nor wildly speculative?

--------------------------------------------------
OUTPUT FORMAT

Return a JSON object:

{{
  "takeaway_comparison": {{
    "interdisciplinary_insightfulness": {{
      "preferred_method": 1 | 2,
      "reasoning": "1-2 sentences explaining your reasoning for the preferred method"
    }},
    "interdisciplinary_relevance": {{
      "preferred_method": 1 | 2,
      "reasoning": "1-2 sentences explaining your reasoning for the preferred method based 
      on the evaluation criteria"
    }},
  }},
  "overall_assessment": {{
    "preferred_method": 1 | 2,
    "summary": "2-3 sentences explaining which method's takeaways are higher quality in 
    terms of interdisciplinary insightfulness and interdisciplinary relevance"
  }}
}}
\end{Verbatim}
\end{tcolorbox}

\begin{tcolorbox}[
    colback=gray!5!white,
    colframe=gray!75!black,
    title={\textbf{Idea Evaluation Prompt}},
    left=2mm,
    right=2mm,
    top=2mm,
    bottom=2mm,
    breakable,
    enhanced
]

\begin{Verbatim}[fontsize=\footnotesize, breaklines=true,breakanywhere=true]
  You are an expert evaluator assessing the quality of cross-domain RESEARCH IDEAS.
Your task is to compare two proposed ideas that integrate insights from an external domain
to address the same research problem.

--------------------------------------------------
RESEARCH PROBLEM
{research_problem}

TARGET DOMAIN
{target_domain}

--------------------------------------------------
METHOD 1 IDEA

Source Domain:
{method_1_idea.get("source_domain", "N/A")}

Proposed Approach:
{method_1_idea.get("idea", {}).get("proposed_approach", "N/A")}

Key Innovations:
{method_1_idea.get("idea", {}).get("key_innovations", [])}

Supporting Takeaways:
{method_1_text}

--------------------------------------------------
METHOD 2 IDEA

Source Domain:
{method_2_idea.get("source_domain", "N/A")}

Proposed Approach:
{method_2_idea.get("idea", {}).get("proposed_approach", "N/A")}

Key Innovations:
{method_2_idea.get("idea", {}).get("key_innovations", [])}

Supporting Takeaways:
{method_2_text}

--------------------------------------------------
EVALUATION CRITERIA

### 1.INTERDISCIPLINARY NOVELTY
Which idea is more novel?
  - The **source domain** chosen and its conceptual distance from the target domain
  - The **proposed_approach**: Is the idea non-obvious to target-domain experts?
  - The **key_innovations**: Do they reflect insights unlikely to arise within the 
  target domain alone?
  - Whether the supporting takeaways draw on **less common or underexplored external 
  insights**

Higher novelty means:
  - The idea is surprising but still credible
  - The cross-domain move feels inventive rather than expected

### 2.INTERDISCIPLINARY USEFULNESS
Which idea has greater interdisciplinary potential for addressing the research problem in 
the target domain ({target_domain})?
  - Ideas with greater interdisciplinary potential should:
    - Present new approaches/solutions to the research problem in the target domain 
    ({target_domain})
    - Address a gap/challenge for the research problem in the target domain 
    ({target_domain})
    - The idea integrates the concepts from both the target domain and source domain 
    into a well-formed idea that addresses the research problem
  - The complexity, simplicity, or practicality of the proposed idea should not factor 
  into your decision (e.g., a more "clear, immediately applicable"/"direct"/"concrete" 
  solution does not make it more useful).Usefulness is defined based on the potential 
  impact of the source domain being introduced to the target domain. Specifically, a more 
  useful interdisciplinary idea integrates the source and target domains in a way that 
  allows for a more significant problem/challenge to be solved or a significant gap in 
  existing ideas to be addressed.
  - Keep in mind that if the distance between the source and target domain is larger 
  (e.g., Computer Science & Engineering are closer than Computer Science & Philosophy), 
  the idea may inherently be less practical. This does not mean that it is less useful. 
  Focus on the degree of the potential impact instead.

--------------------------------------------------
OUTPUT FORMAT

Return a JSON object:

{{
  "idea_comparison": {{
    "interdisciplinary_novelty": {{
      "preferred_method": 1 | 2,
      "reasoning": "1-2 sentences explaining which idea is more novel"
    }},
    "interdisciplinary_usefulness": {{
      "preferred_method": 1 | 2,
      "reasoning": "1-2 sentences explaining why the preferred idea is more useful than 
      the other idea for the research problem based on the evaluation criteria"
    }}
  }},
  "overall_assessment": {{
    "preferred_method": 1 | 2,
    "summary": "2-3 sentences summarizing which idea is overall more interdisciplinary 
    novel, interdisciplinary useful, and integrates the two domains better"
  }}
}}
\end{Verbatim}
\end{tcolorbox}

\end{document}